# Fake Comment Detection Based on Sentiment Analysis

Su Chang,[1] Xu Zhenzhong,[2] Gao Xuan,[3]

*1School of Foreign Languages, Harbin Institute of Technology, Harbin 150001, PR China*
*2College of Computer Science and Technology, Beihang University, Beijing 100191, PR China*
*3State Key Laboratory of Superhard Materials, Jilin University, Changchun 130000, PR China*

**Abstract**

With the development of the E-commerce and reviews website, the comment information is influencing people's life. More and more users share their consumption experience and evaluate the quality of commodity by comment. When people make a decision, they will refer these comments. The dependency of the comments make the fake comment appear. The fake comment is that for profit and other bad motivation, business fabricate untrue consumption experience and they preach or slander some products. The fake comment is easy to mislead users' opinion and decision. The accuracy of humans identifying fake comment is low. It's meaningful to detect fake comment using natural language processing technology for people getting true comment information. This paper uses the sentimental analysis to detect fake comment.

1. **Introduction**

In this Internet time, users can make information which become the comments. The scale of these comments is becoming much bigger. As of 2016, the reviews website, Yelp, has more 108 million comments and the annual growth of the comments is more than 18 million [1]. The comments play a guided role for users' opinion and their consuming behavior. The related statistics show that about 81% users refer the comments. And 80% users think that the comments influence their purchase behavior. Facing the great profit, Internet appear many fake comments[2]. The fake comments are that some users preach and slander goods to influence other users' purchase behavior and opinion. How to identify these fake comments becomes a urgent problem of cybersecurity to solve. It has appeared some ways to detect fake comments. Such as semantic similarity and grammatical analysis. Although these ways can detect fake comments, the accuracy is low. In this paper, we put forward a new method to detect fake comment based on sentimental analysis. As we know, the fake comments often have positive sentiments or negative sentiments. If users release fake comments, they will type many words to express their love and hate. So we use the package, textblob, to evaluate the comments' sentiment. By evaluating, we give every comment a score. We can calculate the distance of the comments by score difference. Then we use the CFSFDP clustering algorithm to identify the fake comments.

The innovation of this paper is as follows:
(1) We first use the sentimental analysis in identifying fake comments. Compared with the previous ways, using sentimental analysis is more suitable for situation.
(2) To identify the fake comments, this paper first uses clustering algorithm. It can gather the comments which have same features together well.

## 2. Data processing

This paper uses the textual data, from Amazon 1930 data [3] about food and commodity, such as follows:

Table1 Raw data

| No | Content |
|---|---|
| 1 | This is a confection that has been around a few centuries.  It is a light, pillow citrus gelatin with nuts - in this case Filberts. And it is cut into tiny squares and then liberally coated with powdered sugar.  And it is a tiny mouthful of heaven.  Not too chewy, and very flavorful.  I highly recommend this yummy treat.  If you are familiar with the story of C.S. Lewis\' "The Lion, The Witch, and The Wardrobe" - this is the treat that seduces Edmund into selling out his Brother and Sisters to the Witch. |
| 2 | his is great and easy to use, but we'll be sticking to peanut butter (or the Kong Stuff'n breath paste). My dog likes it but it seems like such a waste of money when peanut butter works just as well. If you really like kong stuff'n, try the breath paste, it is great! |
| 3 | If you are looking for the secret ingredient in Robitussin I believe I have found it.  I got this in addition to the Root Beer Extract I ordered (which was good) and made some cherry soda.  The flavor is very medicinal |
| 4 | Great taffy at a great price.  There was a wide assortment of yummy taffy.  Delivery was very quick.  If your a taffy lover, this is a deal |
| 5 | I got a wild hair for taffy and ordered this five pound bag. The taffy was all very enjoyable with many flavors: watermelon, root beer, melon, peppermint, grape, etc. My only complaint is there was a bit too much red/black licorice-flavored pieces (just not my particular favorites). Between me, my kids, and my husband, this lasted only two weeks! I would recommend this brand of taffy -- it was a delightful treat. |
| 6 | This saltwater taffy had great flavors and was very soft and chewy.  Each candy was individually wrapped well.  None of the candies were stuck together, which did happen in the expensive version, Fralinger's.  Would highly recommend this candy!  I served it at a beach-themed party and everyone loved it! |
| 7 | This taffy is so good.  It is very soft and chewy.  The flavors are amazing.  I would definitely recommend you buying it.  Very satisfying! |
| 8 | Right now I'm mostly just sprouting this so my cats can eat the grass. They love it. I rotate it around with Wheatgrass and Rye too |
| 9 | This is a very healthy dog food. Good for their digestion. Also good for small puppies. My dog eats her required amount at every feeding. |
| 10 | I don't know if it's the cactus or the tequila or just the unique combination of ingredients, but the flavour of this hot sauce makes it one of a kind!  We picked up a bottle once on a trip we were on and brought it back home with us and were totally blown away!  When we realized that we simply couldn't find it anywhere in our city we were bummed. Now, because of the magic of the internet, we have a case of the sauce and are ecstatic because of it. If you love hot sauce.I mean really love hot sauce, but don't want a sauce that tastelessly burns your throat, grab a bottle of Tequila Picante Gourmet de Inclan.  Just realize that once you taste it, you will never want to use any other sauce. Thank you for the personal, incredible service! |

| 11 | One of my boys needed to lose some weight and the other didn't. I put this food on the floor for the chubby guy, and the protein-rich, no by-product food up higher where only my skinny boy can jump. The higher food sits going stale. They both really go for this food. And my chubby boy has been losing about an ounce a week. |
|---|---|
| 12 | My cats have been happily eating Felidae Platinum for more than two years. I just got a new bag and the shape of the food is different. They tried the new food when I first put it in their bowls and now the bowls sit full and the kitties will not touch the food. I've noticed similar reviews related to formula changes in the past. Unfortunately, I now need to find a new food that my cats will eat. |
| 13 | good flavor! these came securely packed... they were fresh and delicious! i love these Twizzlers! |
| 14 | The Strawberry Twizzlers are my guilty pleasure - yummy. Six pounds will be around for a while with my son and I |
| 15 | My daughter loves twizzlers and this shipment of six pounds really hit the spot. It's exactly what you would expect...six packages of strawberry twizzlers |

TextBlob[4] is *a Pyth*on (2 and 3) library for processing textual data. It provides a simple API for diving into common natural language processing (NLP) tasks such as part-of-speech tagging, noun phrase extraction, sentiment analysis, classification, translation, and more.

The textblob can also analysis the sentiment. Such as ''I fell glad, I feel sad." Textblob first return the classification "I feel glad" and "I feel sad". Then it can calculate the sentiment index. The previous sentence is "polarity=0.8, subjectivity=1.0" and the next sentence is "polarity=-0.5, subjectivity=1.0". The scale of the score is -1 to 1. We can calculate the sum of score. We get the difference of two different scores of two comments.

**3. The CFSFDP clustering algorithm**

"clustering by fast search and find of density peaks"(CFSFDP) [5] is put forward in science by Alex
Rodriguez and Alessandro Laio. This clustering algorithm is novel and simple. The core innovation of the clustering algorithm is how to chose clustering center.

The clustering center has two features:

(1) The clustering has high density which has many points around it.
(2) The distance of different clustering center is far.

The data set for clustering $S = \{x_i\}_{i=1}^{N}, I_s = \{1,2,...,N\}$ is corresponding indicator set. $d_{ij} = dist\ (x_i, x_j)$ is the distance between the data point $x_i, x_j$. For every point $x_i$ of S, we can definite $\rho_i$ and $\delta_i$. These two parameters are to describe the two features.

3.1 $\rho_i$

The following gives the local density of the two formulas.

Gaussian kernel

$$\rho_i = \sum_{j \in I_{s\setminus\{i\}}} e^{-\left(\frac{d_{ij}}{d_c}\right)^2} \quad (1)$$

Cut-off kernel:

$$\rho_i = \sum_{j \in I_{s\setminus\{i\}}} \chi(d_{ij} - d_c) \quad (2)$$

The function:

$$\chi(x) = \begin{cases} 1, & x < 0 \\ 0, & x \geq 0 \end{cases} \quad (3)$$

The parameter, $d_c$, is truncated distance which should be given in advance. According to experience. The $d_c$ should make that the average number of neighbors is around 1 to 2% of the total number of points in the data set. It is very important to choose $d_c$. Because if the value of $d_c$ is large, the distinction of every point density is difficult to identify. In extreme cases, $d_c > d_{max}$, it will lead to only cluster. On the contrary, if the value of $d_c$ is so small, one cluster can divide into many clusters. In the extreme cases, $d_c < d_{min}$, every point is a cluster.

$$d_c = d_{f(Mt)} \quad (4)$$

$f_{(Mt)}$ : Mt to the nearest whole number

For every point of $S = \{x_i\}_{i=1}^{N}, I_s$, it has distances with other N-1 points. So there are N(N-1) distances, but half of them are repeated. Sorting the distance $d_{ij}$ from small to large.( $M = \frac{1}{2}N(N-1)$ points). Assuming the sequence is $d_1 \leq d_2 \leq ... \leq d_M$. If $d_c$ is set as $d_k$, under the condition less than $d_c$, there are $\frac{k}{M}N(N-1)$ distances. For every point, it is about $\frac{k}{M}N(N-1) \approx \frac{k}{M}N$. Obviously, $\frac{k}{M}$ is t. Once t is given, we can get $d_c$.

We can find that Gaussian kernel is continues function and Cut-off kernel is discrete function. In this paper, we use the Cut-off kernel.

3.2 $\delta_i$

$\{q_i\}_{i=1}^{N}$ is the sequence descending order of subscripts. It satisfies that:

$$\rho_{q1} \geq \rho_{q2} \geq ... \geq \rho_{qN} \quad (5)$$

We can definite:

$$\delta_i = \begin{cases} \min_{q_j, j<i}\{d_{q_i}, d_{q_j}\}, i \geq 2 \\ \max_{j \geq 2}\{d_{q_i}, d_{q_j}\}, i = 1 \end{cases} \quad (6)$$

From formula(6), if $x_i$ has the maximum local density, $\delta_i$ means the distance between $x_i$ and other point which has the largest distance to $x_i$. On the contrary, $\delta_i$ means the smallest distance between $x_i$ and other points which density is more than $x_i$.

There are some problems in some situations. For example, if $\rho_i = \rho_j = \max_{j \in I_s}^{\{d_{jk}\}}$, and data point $x_i$, $x_j$ belong to one cluster. The distance between these two points is close. According formula(6), we get that $\delta_i = \max_{j \in I_s}^{\{d_{ij}\}}$ and $\delta_j = \max_{k \in I_s}^{\{d_{jk}\}}$. So if $\delta_i$ and $\delta_j$ are bigger than other points, these two points will be chosen as cluster center that will divide one cluster into two clusters. In order to avoid this question, after sorting, these data points of which $\rho$ are same, then $x_i$ and $x_j$ have the order. Because of this, data point which ranks in front will be chosen as cluster center.

As so far, for every point of S, we can calculate $(\rho_i, \delta_i), i \in I_s$. Such as figure1(A), there are 28 points, and according $\{(\rho_i, \delta_i)\}_{i=1}^{28}, i \in I_s$, plot these points as figure1(B)($\rho$ as X-axis, $\delta$ as Y-axis).

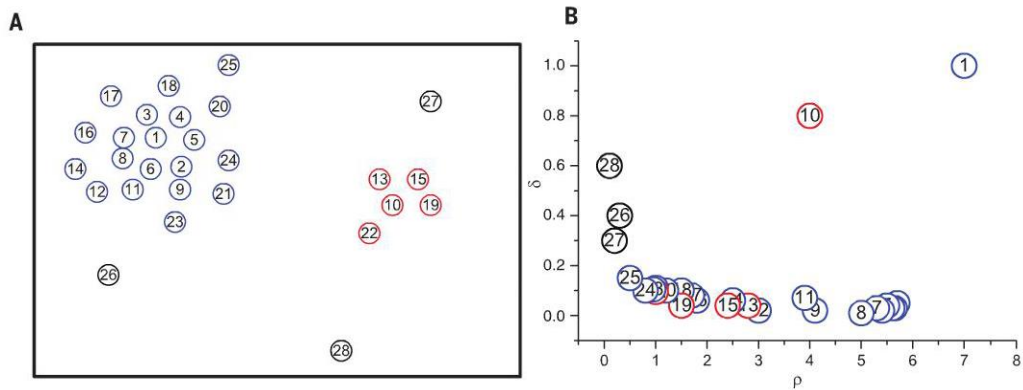

Figure1  Practical examples

It is easy to find that $\rho$ and $\delta$ of number 1 and number 10 are more large, so these two points pop up and they are happened to be the cluster center of figure1(A). Besides, number 26,27,28, these three data points are outlier points. $\delta$ of them is large, but $\rho$ is small.

Figure1(B) is very important to choose cluster center. So we call this figure decision diagram. However, we should notice that we take qualitative analysis to determine cluster center not quantitative analysis. Qualitative analysis includes subjective factors. For different decision diagram, different people can get different opinion.

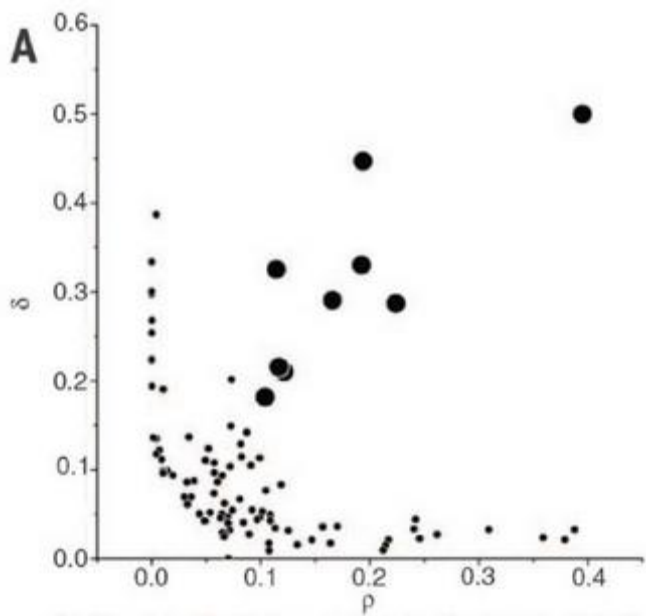

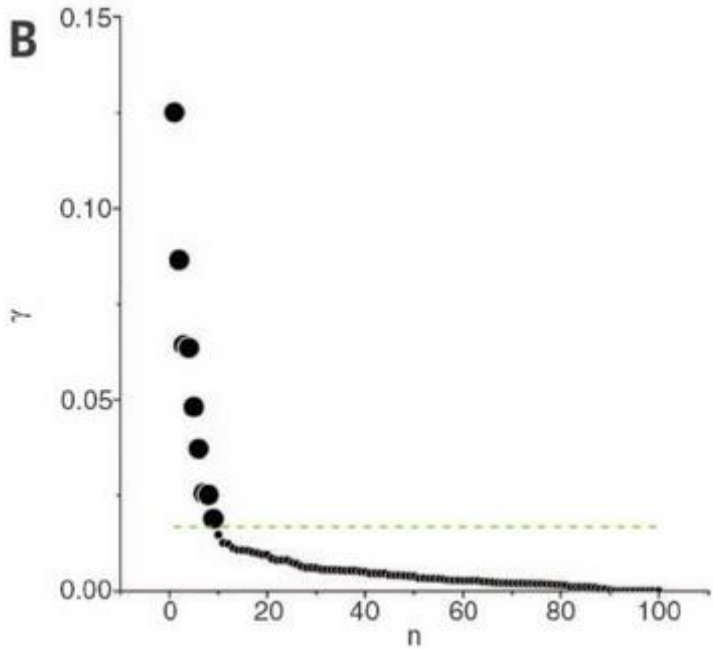

Figure2 Determine cluster center

As figure2(A), it is hard to determine cluster center. Choosing cluster center subjectively ignores the features of data points. For this situation, a parameter which consider $\rho$ and $\delta$ is put forward.

$$\gamma_i = \rho_i \delta_i, i \in I_S \quad (7)$$

According formula(7), we plot as figure2(B). As we can see, the $\gamma$ of data points which are not cluster center is smooth. When the transition to cluster, the curve has a big jump. So we can choose cluster center from this.

When classifying these data points, we depend on the distance of data points to cluster center. Data points choose the cluster center which has smaller distance.

4. **Instance verification**

Through data procession, we get the score of every sentence, and then we get the difference of different sentences.

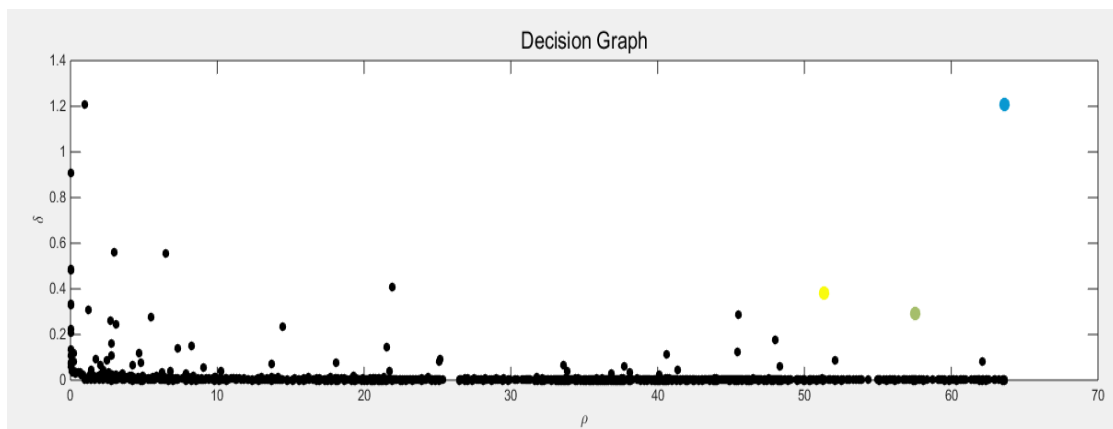

Figure 3 Decision graph

From this graph, we can get three cluster center. They are No 313,603 and 682 data point. These three sentences are:

313: Use frequently as we like to do Asian dishes at least once a week.  Love this product.  Fast shipping, as usual.  Would buy again.

603: Kettle chips are thicker and crunchier (though greasier at times) than other chips, and the honey flavor is great.  They don't have the overwhelming salty taste some cheaper chips have, and the 2oz size is great for an adult size snack.

682: I think Plocky's brand as a whole is a cut above in quality but I found these to lack in taste somewhat and be a little greasy

The classification results are as follows:

Table 3 classification results

| No | Cluster center | amount |
| --- | --- | --- |
| 1 | 313 | 441 |
| 2 | 603 | 1211 |
| 3 | 682 | 278 |

Listing the clustering centers to which the top 20 comments belong:

Table 4 statistical results

| No | Cluster center |
| --- | --- |
| 1 | 603 |
| 2 | 313 |
| 3 | 603 |
| 4 | 603 |
| 5 | 603 |
| 6 | 603 |
| 7 | 603 |
| 8 | 603 |
| 9 | 603 |
| 10 | 313 |
| 11 | 603 |
| 12 | 603 |
| 13 | 682 |
| 14 | 313 |
| 15 | 603 |
| 16 | 603 |
| 17 | 313 |
| 18 | 313 |
| 19 | 603 |
| 20 | 313 |

It is easy to find that No 603 is may be fake comments. In this cluster, there are about 441 points. To ensure the accuracy of the results, we perform iterative clustering. From table 4, we can find which points may be fake comments.

**5. Conclusion**

In this paper, we get the distance between different comments through emotional analysis.

Using CFSFDP to identify fake comments. Through the results, we find that this way can identify fake comments well.

Once we identify fake comments. We can well remove the fake comments. Leaving the correct information is a good thing for consumer.